%% file: egpaper_final.tex
\documentclass[10pt,twocolumn,letterpaper]{article}
\usepackage{titling}
\usepackage{iccv}
\usepackage{times}
\usepackage{epsfig}
\usepackage{graphicx}
\usepackage{amsmath}
\usepackage{amssymb}
\usepackage{textcomp,booktabs}
\usepackage{bbding}
\usepackage{authblk}
\usepackage{footmisc} 
\usepackage[accsupp]{axessibility} 
\usepackage{pdfpages}
\usepackage[title]{appendix}
\usepackage{flushend,cuted}
\usepackage[utf8]{inputenc}
\usepackage{multirow}

\usepackage{placeins}

\newcommand\blfootnote[1]{%
	\begingroup 
	\renewcommand\thefootnote{}\footnote{#1}%
	\addtocounter{footnote}{-1}%
	\endgroup 
}

\usepackage[pagebackref=true,breaklinks=true,letterpaper=true,colorlinks,bookmarks=false]{hyperref}

\iccvfinalcopy 

\def\iccvPaperID{2032} 

\ificcvfinal\fi

\begin{document}
	
	\title{GraphFPN: Graph Feature Pyramid Network for Object Detection}
	
\author[1,2,3]{Gangming Zhao $^{\dagger}$}
\author[1,2]{Weifeng Ge $^*$}
\author[3]{Yizhou Yu $^*$}

\affil[1]{Nebula AI Group, School of Computer Science, Fudan University} 
\affil[2]{Shanghai Key Lab of Intelligent Information Processing} 
\affil[3]{Department of Computer Science, The University of Hong Kong}

	\maketitle
	\ificcvfinal\thispagestyle{empty}\fi
	\blfootnote{{$\dagger$} This work is done when Gangming Zhao is a visiting student at Fudan University. *Corresponding authors: wfge@fudan.edu.cn and yizhouy@acm.org}
	\begin{abstract}
		Feature pyramids have been proven powerful in image understanding tasks that require multi-scale features. State-of-the-art methods for multi-scale feature learning focus on performing feature interactions across space and scales using neural networks with a fixed topology. In this paper, we propose graph feature pyramid networks that are capable of adapting their topological structures to varying intrinsic image structures, and supporting simultaneous feature interactions across all scales. We first define an image specific superpixel hierarchy for each input image to represent its intrinsic image structures. The graph feature pyramid network inherits its structure from this superpixel hierarchy. Contextual and hierarchical layers are designed to achieve feature interactions within the same scale and across different scales. 
		To make these layers more powerful, we introduce two types of local channel attention for graph neural networks by generalizing global channel attention for convolutional neural networks. The proposed graph feature pyramid network can enhance the multiscale features from a convolutional feature pyramid network. 
		
		We evaluate our graph feature pyramid network in the object detection task by integrating it into the Faster R-CNN algorithm. The modified algorithm outperforms not only  previous state-of-the-art feature pyramid based methods with a clear margin but also other popular detection methods on both MS-COCO 2017 validation and test datasets. Codes are available at \tiny{ \url{ https://github.com/GangmingZhao/GraphFPN-Graph-Feature-Pyramid-Network-for-Object-Detection}.}
	\end{abstract}

	\section{Introduction}
	Deep convolutional neural networks exploit local connectivity and weights sharing, and have led to a series of breakthroughs in computer vision tasks, including image recognition~\cite{krizhevsky2012imagenet,szegedy2015going,he2016deep,2019EfficientNet}, object detection~\cite{girshick2014rich, ren2016faster, liu2016ssd, redmon2016you, carion2020end,2017Focal,2020Sparse}, and semantic segmentation~\cite{liu2018path,zhang2018exfuse,lin2019zigzagnet,jha2020doubleu,zhang2020feature,tao2020hierarchical}. 
	Since objects in an image may have varying scales, it is much desired to obtain multiscale feature maps that have fused high-level and low-level features with sufficient spatial resolution at every distinct scale. This motivated feature pyramid networks (FPN~\cite{lin2017feature}) and its improved versions, such as path aggregation network (PANet~\cite{liu2018path}) and feature pyramid transformer (FPT~\cite{zhang2020feature}), and other mtehods~\cite{2018Deep, 2018SAN,2019NAS,2020Auto,2020AugFPN}.
	
	
	Every image has multiscale intrinsic structures, including the grouping of pixels into object parts, the further grouping of parts into objects as well as the spatial layout of objects in the image space. Such multiscale intrinsic structures are different from image to image, and can provide important clues for image understanding and object recognition. But FPN and its related methods always use a fixed multiscale network topology (i.e. 2D grids of neurons) independent of the intrinsic image structures. Such a fixed network topology may not be optimal for multiscale feature learning. According to psychological evidence~\cite{hinton1979some}, human parse visual scenes into part-whole hierarchies, and model part-whole relationships in different images dynamically. Motivated by this, researchers have developed a series of "capsule" models~\cite{sabour2017dynamic,hinton2018matrix,kosiorek2019stacked}, that describe the occurrence of a particular type in a particular region of an image. Hierarchical segmentation can recursively group superpixels according to their locations and similarities to generate a superpixel hierarchy~\cite{pont2016multiscale,maninis2017convolutional}. Such a part-whole hierarchy can assist object detection and semantic segmentation by bridging the semantic gap between pixels and objects~\cite{maninis2017convolutional} .
	
	It is known that multiscale features in a feature pyramid can be enhanced through cross-scale interactions~\cite{lin2017feature,liu2018path,li2019scale,zhang2020feature} in addition to interactions within the same scale. Another limitation of existing methods related to feature pyramid networks is that only features from adjacent scales interact directly while features from non-adjacent scales interact indirectly through other intermediate scales. This is partly because it is most convenient to match the resolutions of two adjacent scales, and partly because it is most convenient for existing interaction mechanisms to handle two scales at a time. Interactions between adjacent scales usually follow a top-down or bottom-up sequential order. In the existing schemes, the highest-level features at the top of the pyramid need to propagate through multiple intermediate scales and interact with the features at these scales before reaching the features at the bottom of the pyramid. During such propagation and interaction, essential feature information may be lost or weakened.
	
	In this paper, we propose graph feature pyramid networks to overcome the aforementioned limitations because graph networks are capable of adapting their topological structures to varying intrinsic structures of input images, and they also support simultaneous feature interactions across all scales. We first define a superpixel hierarchy for an input image. This superpixel hierarchy has a number of levels, each of which consists of a set of nonoverlapping superpixels defining a segmentation of the input image. The segmentations at all levels of the hierarchy are extracted from the same hierarchical segmentation of the input image. Thus the superpixels at two adjacent levels of the hierarchy are closely related. Every superpixel on the coarser level is a union of superpixels on the finer level. Such one-to-many correspondences between superpixels on two levels define the aforementioned part-whole relationships, which can also be called ancestor-descendant relationships. The hierarchical segmentation and the superpixel hierarchy derived from it reveal intrinsic image structures. Although superpixels oversegment an image, pixels in the same superpixel typically belong to the same semantic object/part, and do not straddle the boundaries of semantic objects/parts. Thus, superpixls have more homogeneous pixels than cells from a uniform image partition, and more effectively prevent feature mixing between background clutters and foreground objects.
	
	To effectively exploit intrinsic image structures, the actual structure of our graph feature pyramid network is determined on the fly by the above superpixel hierarchy of the input image. In fact, the graph feature pyramid network inherits its structure from the superpixel hierarchy by mapping superpixels to graph nodes. Graph edges are set up between neighboring superpixels in the same level as well as corresponding superpixels in ancestor-descendant relationships. Correspondences are also set up between the levels in our graph feature pyramid network and a subset of layers in the feature extraction backbone. Initial features at all graph nodes are first mapped from the features at their corresponding positions in the backbone.Contextual and hierarchical graph neural network layers are designed to promote feature interactions within the same scale and across different scales, respectively. Hierarchical layers make corresponding features from all different scales interact directly. Final features at all levels of the graph feature pyramid are fused with the features in a conventional feature pyramid network to produce enhanced multi-scale features.

	Our contributions in this paper are summarized below.
	
	$\bullet$ We propose a novel graph feature pyramid network to exploit intrinsic image structures and support simultaneous feature interactions across all scales. This graph feature pyramid network inherits its structure from a superpixel hierarchy of the input image. Contextual and hierarchical layers are designed to promote feature interactions within the same scale and across different scales, respectively.
	
	$\bullet$ We further introduce two types of local channel attention mechanisms for graph neural networks by generalizing existing global channel attention mechanisms for convolutional neural networks.
	
	$\bullet$ Extensive experiments on MS-COCO 2017 validation and test datasets~\cite{lin2014microsoft} demonstrate that our graph feature pyramid network can help achieve clearly better performance than existing state-of-the-art object detection methods no matter they are feature pyramid based or not. The reported ablation studies further verify the effectiveness of the proposed network components.
	
	
	\begin{figure*}[t]
		\centering
		\includegraphics[width=0.9\linewidth]{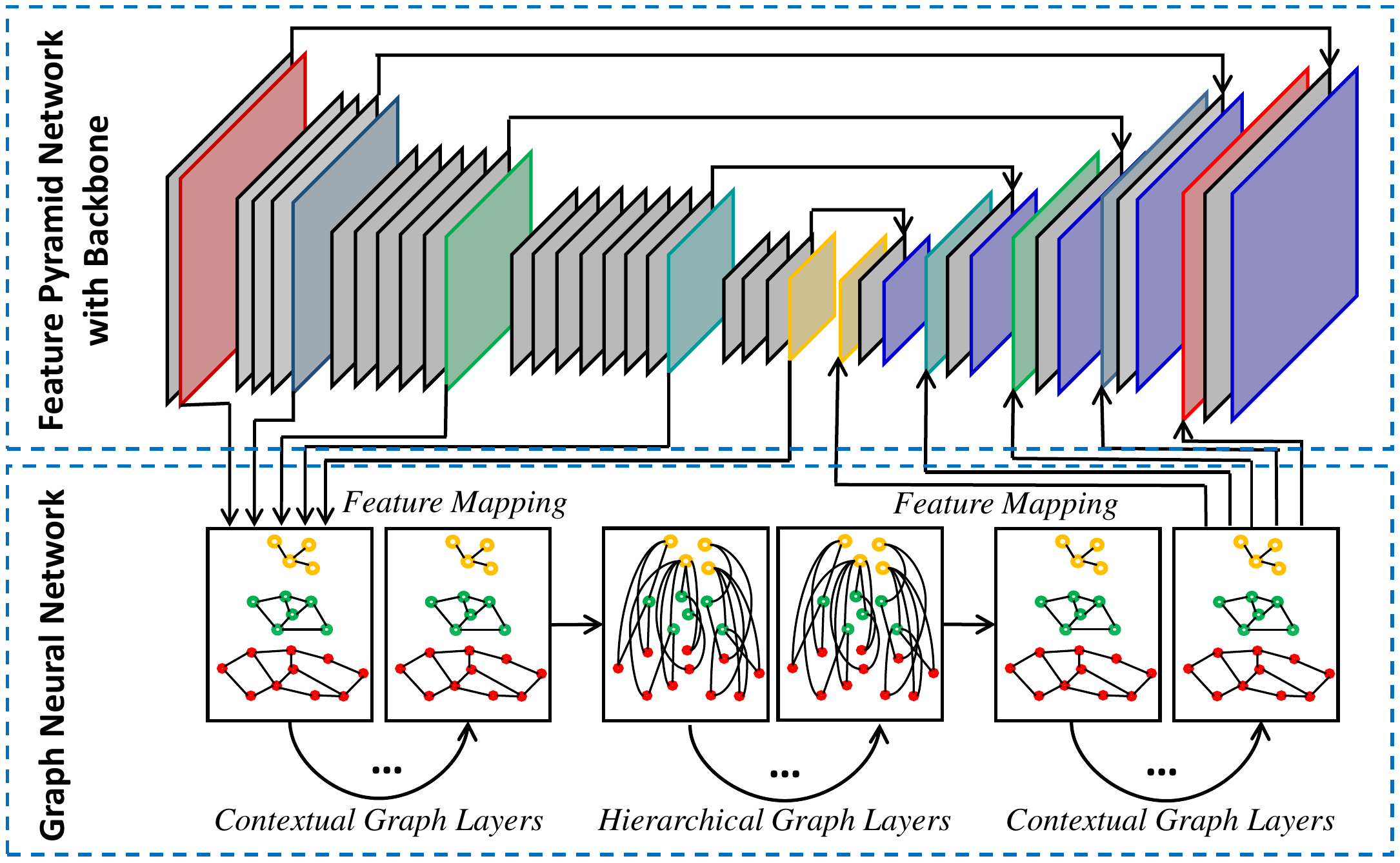}
		\caption{The proposed graph feature pyramid network (GraphFPN) is a graph neural network built on a superpixel hierarchy. GraphFPN receives mapped multi-scale features from the convolutional backbone. These features pass through a number of contextual and hierarchical layers in the GraphFPN before being mapped back to rectangular feature maps, which are then fused with the feature maps from the convolutional FPN for subsequent object detection.}
		\label{Fig:graphfpn}
	\end{figure*}

	\section{Related Work}
	\noindent\textbf{Feature Pyramids.} Feature pyramids present high-level feature maps across a range of scales, and work together with backbone networks to achieve improved and more balanced performance across multiple scales in object detection ~\cite{lin2017feature,liu2018path,li2018detnet,zhao2019m2det,zhang2020feature} and semantic segmentation~\cite{liu2018path,zhang2018exfuse,lin2019zigzagnet,jha2020doubleu,zhang2020feature,tao2020hierarchical}. Recent work on feature pyramids can be categorized into three groups:  top-down networks~\cite{ronneberger2015u,shrivastava2016beyond,lin2017feature,zhang2018exfuse,bilinski2018dense,peng2018megdet}, top-down/bottom-up networks~\cite{lin2018multi,liu2018path}, and attention based methods~\cite{zhang2020feature}. Feature pyramid network (FPN~\cite{lin2017feature}) exploits the inherent multi-scale, pyramidal hierarchy of deep convolutional neural networks, and build a top-down architecture with lateral connections to obtain high level semantic feature maps at all scales. Path Aggregation Network (PANet~\cite{liu2018path}) shortens the information path between lower layers and topmost features with bottom-up path augmentation to enhance the feature hierarchy.
	ZigZagNet~\cite{lin2019zigzagnet} enriches multi-level contextual information not only by dense top-down and bottom-up aggregation, but also by zig-zag crossing between different levels of the top-down and bottom-up hierarchies. Feature pyramid transformer~\cite{zhang2020feature} performs active feature interaction across both space and scales with three transformers. The self-transformer enables non-local interactions within individual feature maps, and the grounding/rendering transformers enable successive top-down/bottom-up interactions between adjacent levels of the feature pyramid. 
	
	In this paper, we aim to fill the semantic gaps between feature maps at different pyramid levels. The most unique characteristic of our graph feature pyramid network in comparison to the above mentioned work is that the topological structure of the graph feature pyramid dynamically adapts to the intrinsic structures of the input image. Furthermore, we build a graph neural network across all scales, making simultaneous feature interactions across all scales possible.
	
	\noindent\textbf{Graph Neural Networks.} Graph neural networks~\cite{li2019deepgcns,velivckovic2017graph,2018Hierarchical,Gong_2019_CVPR,2019Heterogeneous} can model dependencies among nodes flexibly, and can be applied to scenarios with irregular data structures. Graph convolutional networks (GCN~\cite{kipf2016semi}) perform spectral convolutions on graphs to propagate information among nodes. Graph attention networks (GAT~\cite{velivckovic2017graph}) leverage local self-attention layers to designate weights to neighboring nodes, which has gained popularity in many tasks. Gao {\em et al.}~\cite{gao2019graph} proposed graph U-Net with graph pooling and unpooling operations. A graph pooling layer relies on trainable similarity measures to adaptively select a subset of nodes to form a coarser graph while the graph unpooling layer uses saved information to reverse a graph to the structure before its paired pooling operation.
	
	We adopt the self-attention mechanism in GAT~\cite{velivckovic2017graph} in our GraphFPN. To further increase the discriminative power of node features, we introduce local channel attention mechanisms for GNNs by generalizing existing global channel attention mechanisms for CNNs. In comparison to Graph U-Net~\cite{gao2019graph}, our graph pyramid is built on a superpixel hierarchy. Its node merge and split operations are not just based on local similarity ranking, but also depend on intrinsic image structures, which makes our GraphFPN more effective in image understanding tasks.

	\noindent\textbf{Hierarchical Segmentation and GLOM.} Understanding images by building part-whole hierarchies have been a long standing open problem in computer vision~\cite{marr1982vision,bienenstock1997compositionality,belongie2002shape,pantofaru2008object}.
	The hierarchical segmentation algorithms in MCG~\cite{pont2016multiscale} and COB~\cite{maninis2017convolutional} can group pixels of an image into superpixels using detected boundaries. These superpixels are formed hierarchically to describe objects in a bottom-up manner. Hinton~\cite{hinton2021represent} proposed the GLOM imaginary system that aims to use a neural network with a fixed structure to parse images into image specific part-whole hierarchies. 
	
	Given an input image, we use the hierarchical segmentation in COB~\cite{maninis2017convolutional} to build an image specific superpixel hierarchy, on top of which we further build our graph feature pyramid network. One of the contributions of this paper lies in using image specific part-whole hierarchies to enhance multiscale feature learning, which could benefit image understanding tasks including object detection, 
	\begin{figure}[t]
		\centering
		\includegraphics[width=0.7\linewidth]{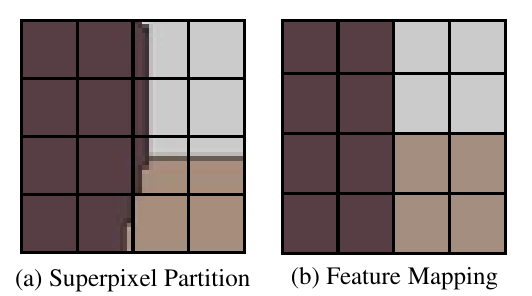}
		\caption{Mapping between CNN grid cells and superpxiels. Each grid cell is assigned to one superpixel it overlaps most. Each superpixel has a small collection of grid cells assigned it.}
		\label{Fig:superpxiel}
	\end{figure}

	\section{Graph Feature Pyramid Networks}
	
	\subsection{Superpixel Hierarchy}
	In a hierarchical segmentation, pixels (or smaller superpixels) are recursively grouped into larger ones with a similarity measure~\cite{pont2016multiscale,maninis2017convolutional}. 
	Given an image $\boldsymbol{I}$, we rely on convolutional oriented boundaries (COB~\cite{maninis2017convolutional}) to obtain a hierarchical segmentation, which is a family of image partitions $\left \{ \mathcal{S}^0,\mathcal{S}^1,...,\mathcal{S}^L \right \}$. Note that each superpixel in $\mathcal{S}^0$ is a single pixel in the original input image, $\mathcal{S}^L$ only has one superpixel representing the entire image, and the number of superpixels in $\mathcal{S}^{l}$ and $\mathcal{S}^{l-1}$ only differ by one (that is, one of the superpixels in $\mathcal{S}^{l}$ is a union of two superpixels in $\mathcal{S}^{l-1}$). 
	
	In this paper, we select a subset of partitions from $\left \{ \mathcal{S}^0,\mathcal{S}^1,...,\mathcal{S}^L\right \}$ to define a superpixel hierarchy $\mathcal{S} = \left\{ \mathcal{S}^{l_1},\mathcal{S}^{l_2},\mathcal{S}^{l_3},\mathcal{S}^{l_4},\mathcal{S}^{l_5}\right \}$, where the superscript of $\mathcal{S}$ stands for the partition level in the segmentation hierarchy, $\mathcal{S}^{l_1}$ is the finest set of superpixels in the hierarchy, and superpixels in $\mathcal{S}^{l_{i+1}}$ are unions of superpixels in $\mathcal{S}^{l_i}$.
	To match the downsampling rate in convolutional neural networks, $\{l_1, l_2, l_3, l_4, l_5 \}$ are chosen such that the number of superpixels in $\mathcal{S}^{l_{i+1}}$ is $1/4$ of that in $\mathcal{S}^{l_i}$. Then the superpixel hierarchy $\mathcal{S}$ can be used to represent the part-whole hierarchy of the input image and track the ancestor-descendant relationships between superpixels.
	
	\subsection{Multi-scale Graph Pyramid}  
	We construct a graph pyramid,  $\left \{\mathcal{G}^1, \mathcal{G}^2, \mathcal{G}^3, \mathcal{G}^4, \mathcal{G}^5 \right \}$, whose levels correspond to levels of the superpixel hierarchy. Every superpixel in the superpixel hierarchy has a corresponding graph node at the corresponding level of the graph pyramid. Thus the number of nodes also decreases by a factor of 4 when we move from one level of the graph pyramid to the next higher level. We define two types of edges for the graph pyramid. They are called {\em contextual edges} and {\em hierarchical edges}. A contextual edge connects two adjacent nodes at the same level while a hierarchical edge connects two nodes at different levels if there is an ancestor-descendant relationship between their corresponding superpixels. Contextual edges are used to propagate contextual information within the same level while hierarchical edges are used for bridging semantic gaps between different levels.
	Note that hierarchical edges are dense because there is such an edge between every node and each of its ancestors and descendants. These dense connections incur a large computational and memory cost. Hence, every hierarchical edge is associated with the cosine similarity between its node features, and we prune hierarchical edges according to their cosine feature similarities. Among all hierarchical edges incident to a node, those ranked in the last 50\% are removed. 
	
	\subsection{Graph Neural Network Layers}
	A graph neural network called GraphFPN is constructed on the basis of the graph pyramid. There are two types of layers in GraphFPN, contextual layers and hierarchical layers. These two types of layers use the same set of nodes in the graph pyramid, but different sets of graph edges. Contextual layers use contextual edges only while hierarchical layers use pruned hierarchical edges only. Our GraphFPN has $L_1$ contextual layers at the beginning, $L_2$ hierarchical layers in the middle and $L_3$ contextual layers at the end. More importantly, each of these layers has its own learnable parameters, which are not shared with any of the other layers. For simplicity, $L_1$, $L_2$ and $L_3$ are always equal in our experiments, and the choice of their specific value is discussed in the ablation studies. The detailed configuration of GraphFPN will be given in the supplementary materials.
	
	Although contextual and hierarchical layers use different edges, GNN operations in these two types of layers are exactly the same. Both types of layers share the same spatial and channel attention mechanisms. We simply adopt the self-attention mechanism in graph attention networks~\cite{velivckovic2017graph} as our spatial attention. Given node ${i}$ and its set of neighbors $\mathcal{N}_i$, the spatial attention updates features as follows, 
	\begin{equation}
		\begin{aligned}
			\vec{h}_i^{\prime}= \boldsymbol{\mathcal{M}}\left ( \vec{h}_i, \left \{ \vec{h}_j \right \}_{j \in \mathcal{N}_i} \right ),
		\end{aligned}
		\label{eq:avgcorr}
	\end{equation}
	where $\boldsymbol{\mathcal{M}}$ is the single-head self-attention from~\cite{velivckovic2017graph}, $\vec{h}_{j \in \mathcal{N}_i}$ is the set of feature vectors collected from the neighbors of node $i$,  $\vec{h}_i$ and $\vec{h}_i^{\prime}$ are respectively the feature vector of node $i$ before and after the update.
	
	The channel attention mechanism is composed of a local channel-wise attention module based on average pooling and a local channel self-attention module. In the average pooling based local channel-wise attention, the feature vectors of node $i$ and its neighbors are first averaged to obtain the feature vector $\vec{a}_i^{\prime} \in \mathbb{R}^C$. We pass the averaged feature vector through a fully connected layer with a sigmoid activation, and perform element-wise multiplication between the result and $\vec{h}_i^{\prime}$, 
	\begin{equation}
		\begin{aligned}
			\vec{h}_i^{\prime \prime} = \sigma (\boldsymbol{W}_1 \vec{a}_i^{\prime}) \odot  \vec{h}_i^{\prime},
		\end{aligned}
		\label{eq:avgcorr}
	\end{equation}
	where $\sigma$ refers to the sigmoid function, $\boldsymbol{W}_1 \in \mathbb{R}^{C \times C}$ is the learnable weight matrix of the fully connected layer, and $\odot$ stands for element-wise multiplication. In the local channel self-attention module, we first obtain the feature vector collection $\boldsymbol{A}$ of node $i$ and its neighbors, and reshape $\boldsymbol{A}$ to $\mathbb{R}^{ \left ( \left | \mathcal{N}_i \right | + 1 \right ) \times C}$. Here $\left | \mathcal{N}_i \right |$ is the size of the neighborhood of node $i$. Next we obtain the channel similarity matrix $\boldsymbol{X} = \boldsymbol{A} ^ \mathrm{ T }  \boldsymbol{A} \in \mathbb{R}^{C \times C}$, and apply the softmax function to every row of $\boldsymbol{X}$. The output of the local channel self-attention module is
	\begin{equation}
		\begin{aligned}
			\vec{h}_i^{\prime \prime \prime} = \beta \boldsymbol{X} \vec{h}_i^{\prime \prime} + \vec{h}_i^{\prime \prime},
		\end{aligned}
		\label{eq:avgcorr}
	\end{equation}   
	where $\beta$ is a learnable weight initialized to 0 as in~\cite{fu2019dual}.
	
	Our local channel-wise attention and local channel self attention are inspired by SENet~\cite{hu2018squeeze} and Dual Attention Network~\cite{fu2019dual}. The main difference is that our channel attention is defined within local neighborhoods and thus spatially varying from node to node while SENet and Dual Attention Network apply the same channel attention to the features at all spatial locations. Advantages of local channel attention in a graph neural network include much lower computational cost and higher spatial adaptivity, and thus is well suited for large networks such as our GraphFPN.
	The ablation study in Table~\ref{tab:messagepassing} demonstrates that our dual local channel attention is rather effective in our GraphFPN.

	\subsection{Feature Mapping between GNN and CNN} 
	Convolutional neural networks can preserve position information of parts and objects, which clearly benefits object detection, while graph neural networks can flexibly model dependencies among parts and objects across multiple semantic scales. Note that the backbone and FPN in a convolutional neural network are respectively responsible for multiscale encoding and decoding while our GraphFPN is primarily responsible for multiscale decoding. Thus features from the backbone serve as the input to the GraphFPN. To take advantage of both types of feature pyramid networks, we also fuse final features from both GraphFPN and convolutional FPN. Therefore, we need to map features from the backbone to initialize the GraphFPN, and also map final features from the GraphFPN to the convolutional FPN before feature fusion. Multi-scale feature maps in the backbone and convolutional FPN are denoted as  $ {\mathcal{C}} = \left \{ {\mathcal{C}}^1,{\mathcal{C}}^2,{\mathcal{C}}^3,{\mathcal{C}}^4,{\mathcal{C}}^5 \right \}$ and  $ {\mathcal{P}} = \left \{ {\mathcal{P}}^1,{\mathcal{P}}^2,{\mathcal{P}}^3,{\mathcal{P}}^4,{\mathcal{P}}^5 \right \}$, respectively. Note that feature maps in ${\mathcal{C}}$ are the final feature maps of the five convolutional stages in the backbone.
	
	\noindent\textbf{Mapping from CNN to GNN ($\mathcal{C} \mapsto \mathcal{S}$):} We map the $i$-th feature map of the backbone $\mathcal{C}^i$ to the $i$-th level $\mathcal{S}^i$ in $\mathcal{S}$. Features in $\mathcal{C}_i$ lie on a rectangular grid, where each grid cell corresponds to a rectangular region in the original input image, while superpixels in $\mathcal{S}^i$ usually have irregular shapes. If multiple superpixels partially overlap with the same grid cell in $\mathcal{C}_i$, as shown in Figure~\ref{Fig:graphfpn}(c), we assign the grid cell to the superpixel with the laregest overlap. Such assignments result in a small collection $C_k^i$ of grid cells assigned to the same superpixel $R_k^i$ in $\mathcal{S}^i$. We perform both max pooling and min pooling over the collection, and feed concatenated pooling results to a fully connected layer with ReLU activation. The mapped feature of $R_k^i$ can be written as
	\begin{equation}
		\begin{aligned}
			\vec{h}_k^i = \delta (\boldsymbol{W}_2 \left [ (\Delta_{max}(C_k^i) \parallel \Delta_{min}(C_k^i)) \right ]),
		\end{aligned}
		\label{eq:avgcorr}
	\end{equation} 
	where $\delta$ stands for the ReLU activation, $\boldsymbol{W}_2$ is the learnable weight matrix of the fully connected layer, $\parallel$ refers to the concatenation operator, and $\Delta_{max}(C_k^i)$ and $\Delta_{min}(C_k^i)$ stand for the max-pooling and min-pooling operators. 

	\begin{table*}[t]\small
		\setlength{\abovecaptionskip}{0pt}
		\setlength{\belowcaptionskip}{0pt}
		\begin{center}
			\resizebox{1.0\linewidth}{!}{
				\begin{tabular}{@{}lcccccccccc@{}}
					\toprule[1pt]
					
					Method           &Training Strategy  &AP & AP$_{50}$ & AP$_{75}$ & AP$_{S}$& AP$_{M}$ &AP$_{L}$ \\ \midrule
					Faster R-CNN~\cite{ren2015faster} &Baseline  & 33.1  & 53.8 & 34.6 & 12.6 & 35.3 & 49.5  \\
					Faster R-CNN+FPN~\cite{lin2017feature} &Baseline  & 36.2 & 59.1 & 39.0 & 18.2 & 39.0 & 52.4  \\
					Faster R-CNN+FPN~\cite{lin2017feature} &MT+AH       & 37.9 & 59.6 & 40.1 & 19.6 & 41.0 & 53.5  \\
					PAN~\cite{liu2018path} &Baseline  & 37.3 & 60.4 & 39.9 & 18.9 & 39.7 & 53.0  \\
					PAN~\cite{liu2018path} &MT+AH     & 39.0 & 60.8 & 41.7 & 20.2 & 41.5 & 54.1  \\
					ZigZagNet~\cite{lin2019zigzagnet} &Baseline  & 39.5  & -- & -- & -- & -- & --  \\	
					ZigZagNet~\cite{lin2019zigzagnet} &MT+AH  & 40.1 & 61.2 & 42.6 & 21.9 & 42.4 & 54.3  \\
					Faster R-CNN+FPN+FPT~\cite{zhang2020feature} &Baseline& 41.6 & 60.9 & 44.0 & 23.4 & 41.5 & 53.1  \\
					Faster R-CNN+FPN+FPT~\cite{zhang2020feature} &AH& 41.1 & 62.0 & 46.6 & 24.2 & 42.1 & 53.3  \\
					Faster R-CNN+FPN+FPT~\cite{zhang2020feature} &MT& 41.2 & 62.1 & 46.0 & 24.1 & 41.9 & 53.2  \\
					Faster R-CNN+FPN+FPT~\cite{zhang2020feature} &MT+AH& 42.6 & 62.4 & 46.9 & 24.9 & 43.0 & \textbf{54.5}  \\   \midrule          	
					Ours &Baseline& 42.1 & 61.3 & 46.1  & 23.6 & 41.1 & 53.3 \\
					Ours &AH& 42.7 & 63.0 & 47.2 & 25.6 & 43.1 & 53.3  \\
					Ours &MT& 42.4 & 62.7 & 46.9 & 24.3 & 43.1 & 53.6  \\
					Ours &MT+AH&\textbf{43.7}($\uparrow$1.1)  & \textbf{64.0}($\uparrow$1.6) & \textbf{48.2}($\uparrow$1.3) & \textbf{27.2}($\uparrow$2.3) & \textbf{43.4}($\uparrow$0.4) & {54.2}($\downarrow$0.3)  \\           	                              	
					\bottomrule[1pt]
			\end{tabular}}
		\end{center}
		\caption{Comparison with state-of-the-art feature pyramid based methods on MS-COCO 2017 test-dev~\cite{lin2014microsoft}. ``AH'' and ``MT'' stand for augmented head and multi-scale training strategies~\cite{liu2018path} respectively. 
			The backbone of all listed methods is ResNet101~\cite{he2016deep}.}
		\label{tab:sota-coco-test}
	\end{table*}
	
	\begin{table*}[t]\small
		\setlength{\abovecaptionskip}{0pt}
		\setlength{\belowcaptionskip}{0pt}
		\begin{center}
			\resizebox{0.95\linewidth}{!}{
				\begin{tabular}{@{}lccccccccc@{}}
					\toprule[1pt]
					
					Method                                  &Detection Framework &AP & AP$_{50}$ & AP$_{75}$ & AP$_{S}$& AP$_{M}$ &AP$_{L}$ \\ \midrule
					RetinaNet + FPN~\cite{2017Focal}        &RetineNet  & 40.4  & 60.2 & 43.2 & 24.0 & 44.3 & 52.2  \\
					Faster R-CNN+FPN~\cite{lin2017feature}  &Faster R-CNN & 42.0 & 62.5 & 45.9 & 25.2 & 45.6 & 54.6  \\
					DETR~\cite{carion2020end}              &Set Prediction & 44.9 & 64.7 & 47.7 & 23.7 & 49.5 & 62.3  \\
					Deformable DETR~\cite{zhu2020deformable} & Set Prediction & 43.8  & 62.6 & 47.7 & 26.4 & 47.1 & 58.0  \\	
					Sparse R-CNN+FPN ~\cite{zhang2020feature} & Sparse R-CNN & 45.6 & 64.6 & 49.5 & 28.3 & 48.3 & 61.6  \\  \midrule          	
					Ours &Faster R-CNN &\textbf{46.7}($\uparrow$1.1)  & \textbf{65.1}($\uparrow$0.5) & \textbf{50.1}($\uparrow$0.6) & \textbf{29.2}($\uparrow$0.9) & 49.1($\downarrow$0.8) & 61.8($\downarrow$0.2)  \\           	                              	
					\bottomrule[1pt]
			\end{tabular}}
		\end{center}
		\caption{Comparison with other popular object detectors on MS-COCO 2017 val set~\cite{lin2014microsoft}. The backbone of all listed methods is ResNet101~\cite{he2016deep}.}
		\label{tab:sota-coco-val}
	\end{table*}


	\noindent\textbf{Mapping from GNN to CNN ($\mathcal{S} \mapsto \mathcal{P}$):}
	Once we run a forward pass through the GraphFPN, we map the features of its last layer to the convolutional feature pyramid $\mathcal{P}$. Let $P_k^i$ be the collection of grid cells in $\mathcal{P}^i$ assigned to the superpixel $R_k^i$ in $\mathcal{S}^i$. We simply copy the final feature at $R_k^i$ to every grid cell in $P_k^i$. In this way, we obtain a new feature map $\overline{\mathcal{P}}^i$ for the $i$-th level of the convolutional FPN. We concatenate $\mathcal{P}^i$ with $\overline{\mathcal{P}}^i$, and feed the concatenated feature map to a convolutional layer with $1 \times 1$ kernels to ensure the fused feature map $\widetilde{\mathcal{P}}^i$ has the same number of channels as $\mathcal{P}^i$. Finally, the fused feature pyramid is $ \widetilde{\mathcal{P}} = \left \{\widetilde{\mathcal{P}}^1, \widetilde{\mathcal{P}}^2,\widetilde{\mathcal{P}}^3,\widetilde{\mathcal{P}}^4,\widetilde{\mathcal{P}}^5 \right \}$.
	
	\subsection{Object Detection}
	The proposed graph feature pyramid network can be integrated into the object detection pipeline in \cite{lin2017feature} by replacing the conventional FPN with the above fused feature pyramid. We adopt faster-RCNN as our detection algorithm, and perform the same end-to-end training. In the following section, we conduct extensive experiments in object detection to validate the effectiveness of the proposed method.

	\section{Experiments}
	\noindent\textbf{Datasets.}
	We evaluate the proposed method on MS COCO 2017 detection dataset~\cite{lin2014microsoft}, containing 118k training images, 5k validation images and 20k testing images. Metrics for performance evaluation include the standard average precision (AP), ${\rm AP}_{50}$, ${\rm AP}_{75}$, ${\rm AP}_{S}$, ${\rm AP}_{M}$, and ${\rm AP}_{L}$. We report ablation study results on the validation set, and report results on the standard test set to compare with state-of-the-art algorithms.
	
	
	\noindent\textbf{Implementation details.}
	We have fully implemented our GraphFPN using PyTorch, and all models used in this paper are trained on 8 NVidia TITAN 2080Ti GPUs.
	As a common practice~\cite{lin2017feature,lin2019zigzagnet}, all backbone networks are pretrained on the ImageNet1k image classification dataset~\cite{krizhevsky2012imagenet}, and then fine-tuned on the training set of the detection dataset. Faster-RCNN~\cite{ren2015faster} is adopted as our object detection framework, and we follow the settings in FPT~\cite{zhang2020feature} to set up the detection heads. During training, we adopt Adam~\cite{kingma2014adam} as our optimizer, and set the weight decay and momentum to 0.0001 and 0.9 respectively. Every mini-batch contains 16 images, and are distributed on 8 GPUs with the synchronized batch norm (SBN~\cite{zhang2018context}). For fair comparison, input images are resized to 800/1,000 pixels along the shorter/longer edge. The models used in all experiments are trained with 36 epochs on the detection training set. The initial learning rate is set to 0.001, which is decreased by a factor of 10 at the $27$-th and $33$-th epochs respectively. It takes 38 hours to train a faster-RCNN model integrated with our GraphFPN on the COCO dataset.
	
	We use codes provided by the COB project\footnote{https://cvlsegmentation.github.io/cob/}~\cite{maninis2017convolutional} to compute hierarchical segmentations, and build a superpixel hierarchy for each image during data preparation. It takes 0.120 seconds on average to build the superpixel hierarchy of an image, which is reasonable for an object detection task. Note that machine learning models used in COB are always trained on the same training set as the detection task.
	\begin{figure*}[t]
		\centering
		\includegraphics[width=0.95\linewidth]{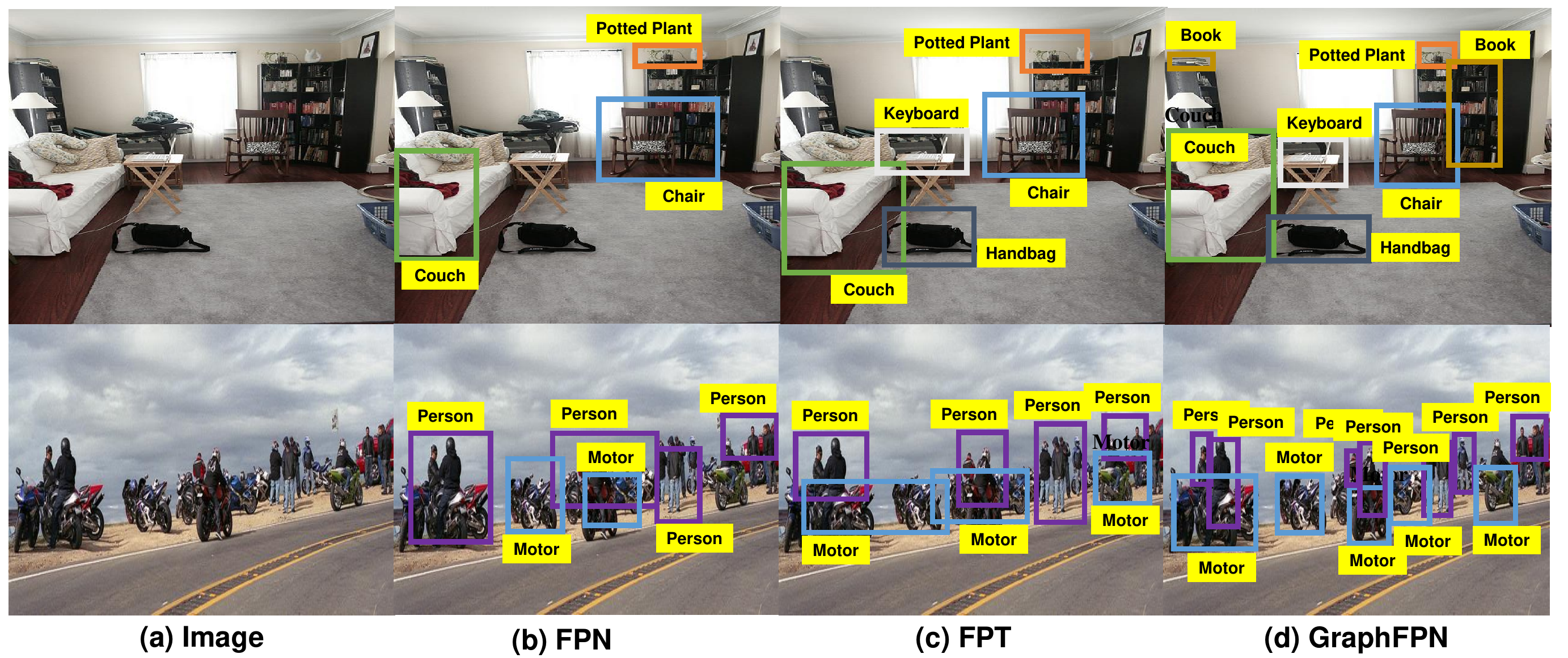}
		\caption{Sample detection results from FPN~\cite{lin2017feature}, FPT~\cite{zhang2020feature}, and our GraphFPN based method. 
		}
		\label{Fig:visual}
	\end{figure*}

	\begin{table*}[h]
		\begin{center}
			\resizebox{0.6\linewidth}{!}{
				\begin{tabular}{l|ccc}
					\toprule[1pt]
					Methods & Params & GFLOPs & Test Speed (s) \\
					\midrule
					faster RCNN~\cite{ren2015faster} & 34.6 M & 172.3 & 0.139  \\
					faster RCNN + FPN~\cite{lin2017feature}  & 64.1 M & 240.6 & 0.051  \\
					faster RCNN + FPN + FPT~\cite{zhang2020feature} & 88.2 M & 346.2 & 0.146 \\
					\midrule
					faster RCNN + FPN + GraphFPN  & 100.0 M & 380.0 & 0.157 \\
					COB + faster RCNN + FPN + GraphFPN  & 121.0 M & 393.1 & 0.277 \\
					\bottomrule[1pt]
				\end{tabular}
			}
		\end{center}
		\caption{The number of learnable parameters, the total computational cost, and the average test speed of a few detection models. All experiments are run on an NVidia TITAN 2080Ti GPU.}\label{tab:complex}
	\end{table*}

	\subsection{Comparison with State-of-the-Art Methods}~\label{exp:sota}
	We compare the object detection performance of our method (GraphFPN+FPN) with existing state-of-the-art feature pyramid based methods, including feature pyramid networks (FPN~\cite{lin2017feature}), path aggregation networks (PANet~\cite{liu2018path}), ZigZagNet~\cite{lin2019zigzagnet} and feature pyramid transformers (FPT~\cite{zhang2020feature}), using Faster-RCNN as the detection framework to verify the effectiveness of feature interactions in both the contextual layers and the hierarchical layers.
	
	Table~\ref{tab:sota-coco-test} shows experimental results achieved with the above mentioned state-of-the-art methods on MS COCO 2017 {\em{test-dev}}~\cite{lin2014microsoft} in various settings. Our method achieves the highest AP (43.7\%) outperforming other state-of-the-art algorithms by at least 1.1\%, and maintains a leading role on AP$_50$, AP$_75$, AP$_S$, and AP$_M$. When compared with the Faster-RCNN baseline~\cite{ren2015faster}, the AP of our method is 10.6\% higher. It indicates that multi-scale high-level feature learning is crucial for object detection. When our method is compared with FPN alone~\cite{lin2017feature}, the improvement in AP reaches 7.5\%, which further indicates that GraphFPN significantly enhances the original multi-scale feature learning conducted with FPN, and multi-scale feature interaction and fusion are very effective for object detection. Such improvements also illustrate that graphs built on top of superpixel hierarchies are capable of capturing intrinsic structures of images, and are helpful in high-level image understanding tasks. 
	When compared with FPT~\cite{zhang2020feature}, our method achieves better performance on five evaluation metrics, including AP, AP$_{50}$, AP$_{75}$, AP$_{S}$ and AP$_{M}$, except AP$_{L}$. We attribute this performance to three factors. First, graph neural networks have higher efficiency in propagating information across different semantic scales by connecting nodes dynamically while FPT has to broadcast information in a cascaded manner with the top-down and bottom-up combination. Second, the superpxiel hierarchy captures the intrinsic structures of images, which benefit the detection of small-scale objects. Then our method achieves 2.3\% improvement on AP$_{S}$ in comparison to FPT. Third, superpixel hierarchies are not well suited for the detection of large-scale objects, which can be verified through the inferior result on AP$_{L}$.

	\subsection{Comparison with Other Object Detectors}
	In addition to comparison with feature pyramid based detection methods, we further compare our method with other popular detectors. As shown in Table~\ref{tab:sota-coco-val}, our method based on Faster R-CNN + FPN + GraphFPN outperforms all such detectors, including RetinaNet~\cite{2017Focal}, DETR~\cite{carion2020end}, Deformable DETR~\cite{zhu2020deformable} and Sparse R-CNN+FPN ~\cite{zhang2020feature}, by a clear margin when they use the same backbone as our method. Our method achieves compelling performance under all six performance metrics. This demonstrates our GraphFPN is capable of significantly enhancing the feature representation of a detection network, which in turn leads to superior detection performance.

	\subsection{Learnable Parameters and Computational Cost}
	Table~\ref{tab:complex} provides the number of learnable parameters, the total computational cost, and the average test speed of a few detection models. 
	Faster RCNN~\cite{ren2015faster} serves as our baseline, which has 34.6 million learnable parameters and 172.3 GFLOPs. It takes 0.139 seconds on average to process one image. Our GraphFPN works on top of Faster RCNN and FPN, and the whole pipeline has 1.89 times more learnable parameters, 1.21 times more GFLOPs and 12.9\% longer test time. 
	If we take the construction of the superpixel hierarchy into consideration, COB~\cite{maninis2017convolutional} models have 21 (+21\%) million parameters, 13.1 (+3.4\%) GFLOPs, and 0.12 (+76.4\%) seconds time cost. This is because COB~\cite{maninis2017convolutional} needs to detect contours in an image and build a hierarchical segmentation on CPU. In fact, hierarchical segmentation could be implemented using CUDA and run on GPU, which would significantly reduce the test time.

	\subsection{Ablation Studies}
	
	\begin{table}[t]
		\begin{center}
			\resizebox{0.9\linewidth}{!}{	
				\begin{tabular}{ccccccc}
					\toprule[1pt]
					CGL-1 & HGL & CGL-2 & AP           & AP$_{S}$ & AP$_{M}$ & AP$_{L}$ \\
					\midrule
					$\checkmark$ & $\checkmark$& $\checkmark$  & \textbf{39.1}  & \textbf{22.4}  & \textbf{38.9} &  \textbf{56.7} \\
					$\times$  & $\checkmark$& $\checkmark$  & 38.2  & 22.1  & 38.7 &  56.1 \\
					$\checkmark$ & $\checkmark$& $\times$   & 38.7  & 22.1  & 38.9 &  56.6 \\
					$\times$  & $\checkmark$& $\times$   & 36.2  & 19.2  & 36.3 &  54.4 \\
					$\checkmark$& $\times$  &  $\times$  & 37.2  & 22.1  & 35.1 &  55.6 \\
					\bottomrule[1pt]
				\end{tabular}
			}
		\end{center}
		\caption{Ablation study on the contextual and hierarchical layers in GraphFPN. ``CGL-1'' stands for the first group of contextual layers before the hierarchical layers, ``HGL'' stands for the hierarchical layers, and ``CGL-2'' stands for the second group of contextual layers after the hierarchical layers. $\checkmark$ and $\times$ stand for the existence of a module or not. Detection results are reported on the MS-COCO 2017 val set~\cite{lin2014microsoft}.}\label{tab:graphupdate}
	\end{table}

	\begin{table}[t]
		\begin{center}
			\resizebox{0.8\linewidth}{!}{
				\begin{tabular}{ccccccc}
					\toprule[1pt]
					SA & LCA & LSA & AP           & AP$_{S}$ & AP$_{M}$ & AP$_{L}$ \\
					\midrule
					$\checkmark$ & $\checkmark$  & $\checkmark$&  \textbf{39.1}  & \textbf{22.4}  & \textbf{38.9} &  \textbf{56.7} \\
					$\times$  & $\checkmark$& $\checkmark$  & 37.8  & 21.9  & 37.4 &  56.2 \\
					$\checkmark$ & $\times$ & $\checkmark$  & 37.9 & 21.6  & 37.3 &  56.4 \\
					$\checkmark$ & $\checkmark$ & $\times$  & 37.6  & 21.8  & 37.7 &  55.1 \\
					$\checkmark$  & $\times$ & $\times$&  37.1  & 21.1  & 36.7 &  54.1 \\
					\bottomrule[1pt]
				\end{tabular}
			}
		\end{center}
		\caption{Ablation study on the attention mechanisms. ``SA'' stands for the spatial attention module, ``LCA'' stands for the local channel-wise attention module  and ``LSA'' stands for the local channel self-attention module. $\checkmark$ and $\times$ stand for the existence of a module or not. Detection results are reported on the MS-COCO 2017 val set~\cite{lin2014microsoft}.}\label{tab:messagepassing}
	\end{table}

	\begin{table}[t]
		\begin{center}		
			\resizebox{0.8\linewidth}{!}{
				\begin{tabular}{ccccccc}
					\toprule[1pt]
					N & AP & AP$_{50}$ & AP$_{75}$ & AP$_{S}$ & AP$_{M}$ & AP$_{L}$ \\
					\midrule
					1 & 36.1 & 56.3 & 35.4 & 19.3  & 37.9 &  55.4 \\
					2  & 37.2 & 57.6 & 38.5 & 21.2  & 38.3 &  55.8 \\
					3  & \textbf{39.1} & \textbf{58.3} & \textbf{39.4} & \textbf{22.4}  & \textbf{38.9} &  \textbf{56.7}  \\
					4  & 38.1 & 57.8 & 38.9 & 22.2  & 38.6 &  56.3  \\
					5  & 37.1 & 57.1 & 38.0 & 21.9  & 37.9 &  55.4  \\
					\bottomrule[1pt]
				\end{tabular}
			}
		\end{center}
		\caption{Ablation study on the number of layers in GraphFPN. N is the number of layers in each of the three groups of layers. Hence, the total number of layers is 3N. Detection results are reported on the MS-COCO 2017 val set~\cite{lin2014microsoft}.}\label{tab:graphlayer}
	\end{table}

	To investigate the effectiveness of individual components in our GraphFPN, we conduct ablation studies by replacing or removing a single component from our pipeline. We have specifically designed ablation studies for the configuration of GNN layers (the combination and ordering of different types of GNN layers), the total number of GNN layers, and the spatial and channel attention mechanisms.
	
	\noindent\textbf{GNN Layer Configuration.}
	In our final pipeline, the specific configuration of layers is as follows: first group of contextual layers, a group of hierarchical layers, and second group of contextual layers. The number of layers in all groups are the same.
	Table~\ref{tab:graphupdate} shows the results of the ablation study on the configuration of these layers. When we remove the first group of contextual layers, the AP drops by 0.9\%. It means that it is necessary to propagate contextual information within the same scale before cross-scale operations. Then we remove the second group contextual layers, the AP drops by 0.4\%, which indicates contextual information propagation is still helpful even after the first group of contextual layers followed by a group of hierarchical layers. If we keep one group of contextual layers or hierarchical layers only, the AP drops by 2.9\% and 1.9\% respectively, which indicates the two types of layers are truly complementary to each other.
	
	\noindent\textbf{Number of GNN Layers.}
	The number of layers in a GNN affects its overall discriminative ability. Table~\ref{tab:graphlayer} shows experimental results with different numbers of layers in each type. When $L=3$, which means each of the three groups has 3 layers and the total number of layers is 9, our method achieves the best results on all five performance metrics. When there are too many graph layers, the performance becomes worse. We attribute this to gradient vanishing.
	
	\noindent\textbf{Attention Mechanism.}
	In the ablation study shown in Table~\ref{tab:messagepassing}, we verify the effectiveness of the spatial self-attention and the two local channel attention mechanisms. When we remove the spatial self-attention, the AP drops by 1.3\%. It means that the spatial attention is powerful in modeling neighborhood dependencies. If we remove the local average-pooling based channel-wise attention or the local channel self-attention, the AP drops by 1.2\% and 1.5\% respectively. It demonstrates that these two local channel attention mechanisms are complementary to each other, and significantly improve the discriminative ability of deep features. If we completely remove both channel attention mechanisms, the AP is 2\% worse.

	\section{Conclusions}
	In this paper, we have presented graph feature pyramid networks that are capable of adapting their topological structures to varying intrinsic structures of input images, and supporting simultaneous feature interactions across all scales. Our graph feature pyramid network inherits its structure from a superpixel hierarchy constructed according to a hierarchical segmentation. Contextual and hierarchical graph neural network layers are defined to achieve feature interactions within the same scale and across different scales, respectively. 
	To make these layers more powerful, we further introduce two types of local channel attention for graph neural networks. Extensive experiments demonstrate that Faster R-CNN+FPN integrated with our graph feature pyramid network outperforms existing state-of-the-art object detection methods on MS-COCO 2017 validation and test datasets.
	{\small
		\bibliographystyle{ieee_fullname}
		\bibliography{egpaper_final}
	}
	
	\clearpage
	\input{supplementary}

\end{document}

%% file: supplementary.tex
\makeatletter
\def\@thanks{}
\makeatother

\renewcommand\arraystretch{2}
\setlength{\affilsep}{0.4em}

\title{\emph{Supplementary Material} for \\ GraphFPN: Graph Feature Pyramid Network for Object Detection}

\makeatletter
\renewcommand{\@maketitle}{
\newpage
\null
\vskip 2em
\begin{center}
    {
    \Large \@title \par
    }
\end{center}
\par}\makeatother

\renewcommand\thesection{\Alph{section}}
\setcounter{section}{0}
\setcounter{table}{0}
\setcounter{figure}{0}

\onecolumn
\maketitle
\setlength{\parindent}{1em}
\setlength{\parskip}{0em}

\section{Network Architecture and Visual Results}

We provide the network architecture of the backbone (ResNet-101~\cite{he2016deep}) and FPN used in the proposed pipeline in Table~\ref{tab:backbone+FPN}. The network architecture of our GraphFPN is given in Table~\ref{tab:graphFPN}. For our GraphFPN, the feature dimension $F$ of every graph node is always set to 256 in all experiments reported in this paper. In GraphFPN, the first group of three layers are contextual layers, the second group of three layers are hierarchical layers, and the last group of three layers are contextual layers again. As mentioned in the paper, the graphs in all these layers have identical sets of nodes (distributed in five levels), but contextual and hierarchical layers have different sets of graph edges. Each of these layers has three attention modules, a spatial self-attention module, a local channel-wise attention module and a local channel self-attention module. Note that the number of graph nodes in each layer of GraphFPN is $(N+\frac{N}{4}+\frac{N}{16}+\frac{N}{64}+\frac{N}{256})$, where $N$ is the number of superpixels in the finest level of a superpixel hierarchy. 

\begin{figure*}[t]
	\centering
	\includegraphics[width=1.0\linewidth]{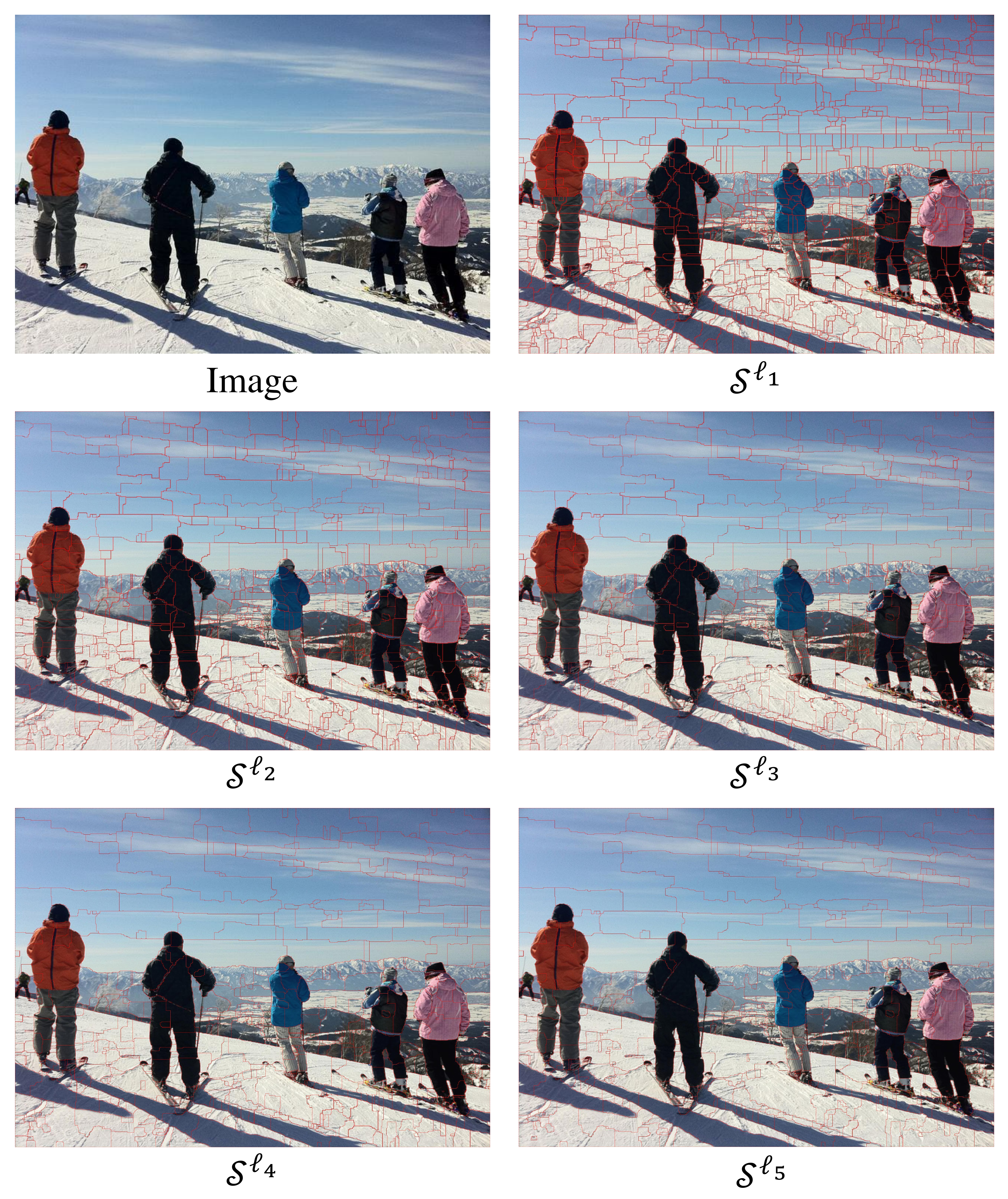}
	\caption{Sample result of superpixel hierarchy. Each superpixel hierarchy consists of 5 levels, $\left\{ \mathcal{S}^{l_1},\mathcal{S}^{l_2},\mathcal{S}^{l_3},\mathcal{S}^{l_4},\mathcal{S}^{l_5}\right \}$. Images are from the MS COCO 2017 dataset~\cite{lin2014microsoft}.}
	\label{Fig:sp-visual1}
\end{figure*}

\begin{figure*}[t]
	\centering
	\includegraphics[width=1.0\linewidth]{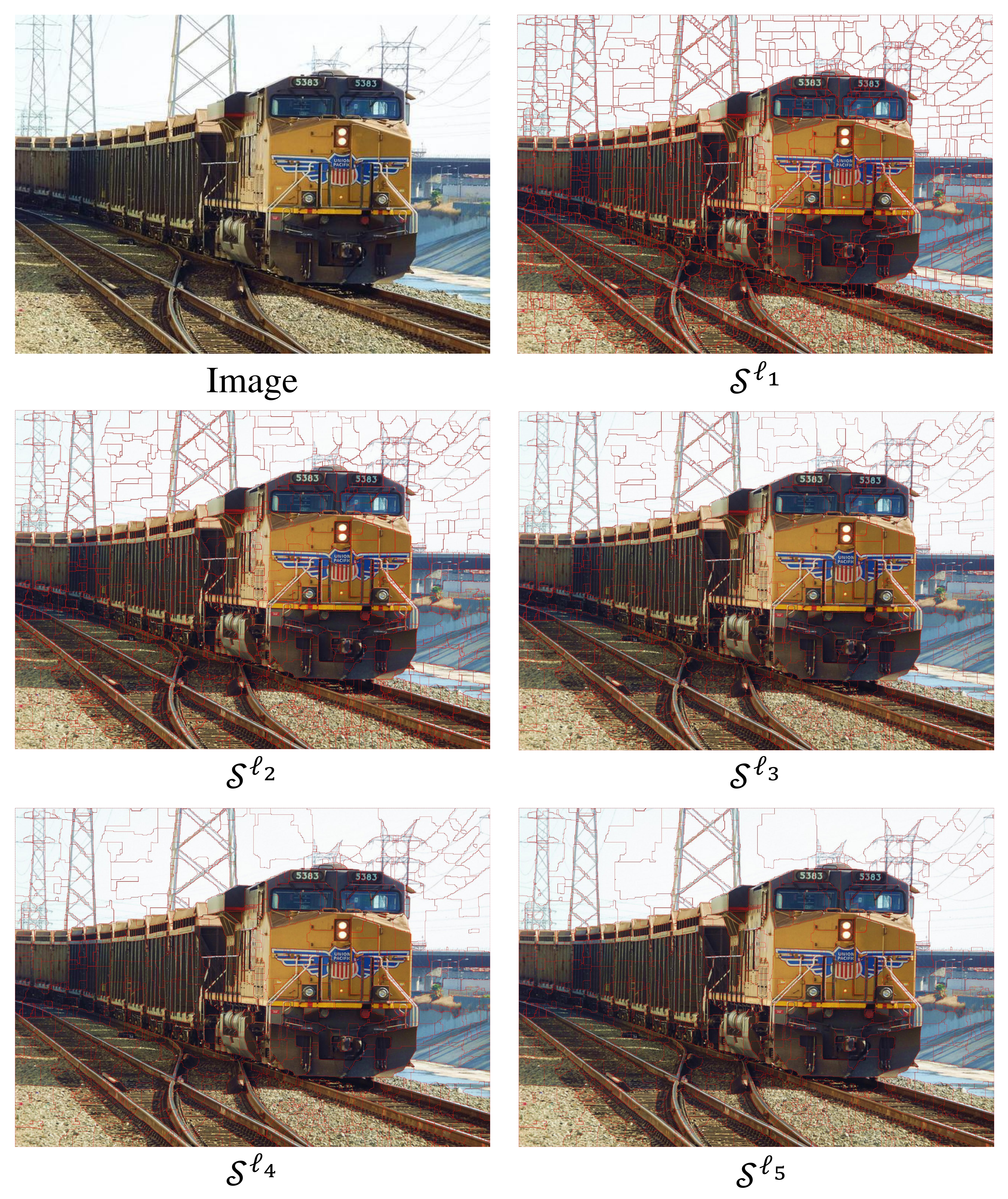}
	\caption{Sample result of superpixel hierarchy. Each superpixel hierarchy consists of 5 levels, $\left\{ \mathcal{S}^{l_1},\mathcal{S}^{l_2},\mathcal{S}^{l_3},\mathcal{S}^{l_4},\mathcal{S}^{l_5}\right \}$. Images are from the MS COCO 2017 dataset~\cite{lin2014microsoft}.}
	\label{Fig:sp-visual2}
\end{figure*}

\begin{figure*}[t]
	\centering
	\includegraphics[width=1.0\linewidth]{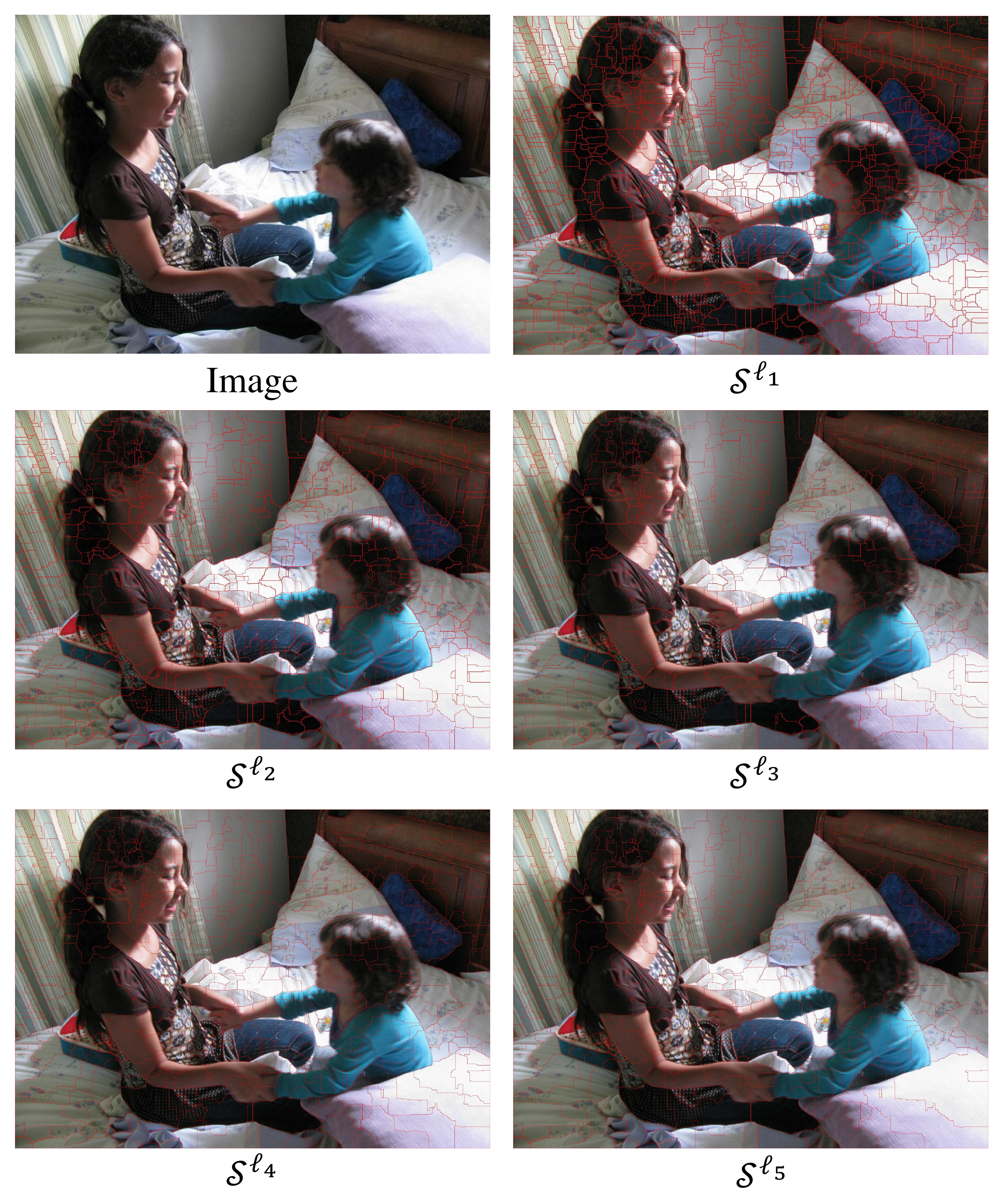}
	\caption{Sample results of superpixel hierarchy. Each superpixel hierarchy consists of 5 levels, $\left\{ \mathcal{S}^{l_1},\mathcal{S}^{l_2},\mathcal{S}^{l_3},\mathcal{S}^{l_4},\mathcal{S}^{l_5}\right \}$. Images are from the MS COCO 2017 dataset~\cite{lin2014microsoft}.}
	\label{Fig:sp-visual3}
\end{figure*}

Figures~\ref{Fig:sp-visual1},~\ref{Fig:sp-visual2}, and~\ref{Fig:sp-visual3} show sample superpixel hierarchies based on hierarchical image segmentation algorithm COB~\cite{maninis2017convolutional}. Starting from the finest partition $\mathcal{S}^{l_1}$, superpixels are recursively merged according to contour strengths to generate a set of partitions and form a superpixel hierarchy $\left\{ \mathcal{S}^{l_1},\mathcal{S}^{l_2},\mathcal{S}^{l_3},\mathcal{S}^{l_4},\mathcal{S}^{l_5}\right \}$. Input images are taken from the MS COCO 2017 dataset~\cite{lin2014microsoft}.

\begin{figure*}[t]
	\centering
	\includegraphics[width=1.0\linewidth]{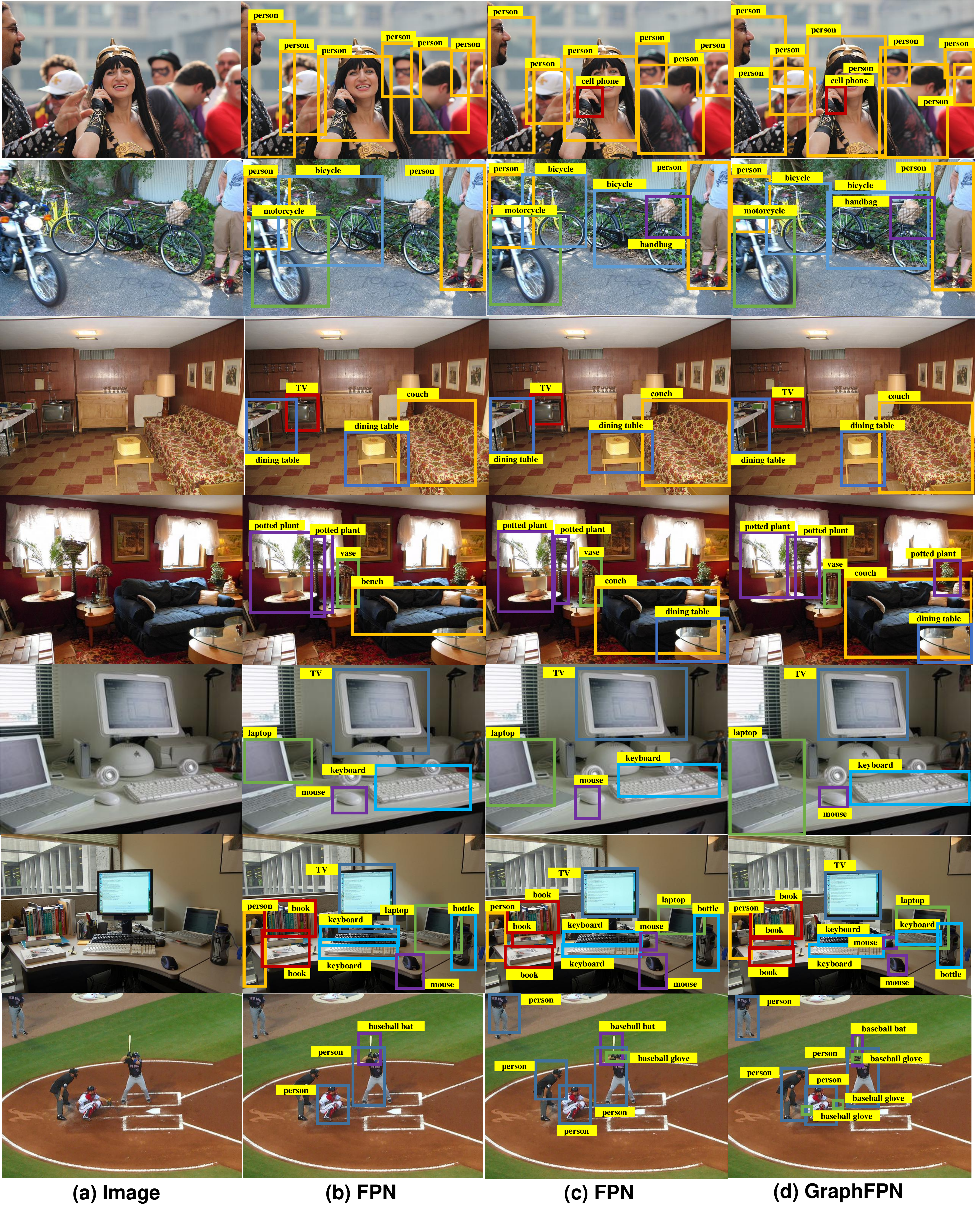}
	\caption{Sample detection results from FPN~\cite{lin2017feature}, FPT~\cite{zhang2020feature}, and our GraphFPN based method. Images are from the MS COCO 2017 validation set~\cite{lin2014microsoft}.}
	\label{Fig:res-visual1}
\end{figure*}

\begin{figure*}[t]
	\centering
	\includegraphics[width=1.0\linewidth]{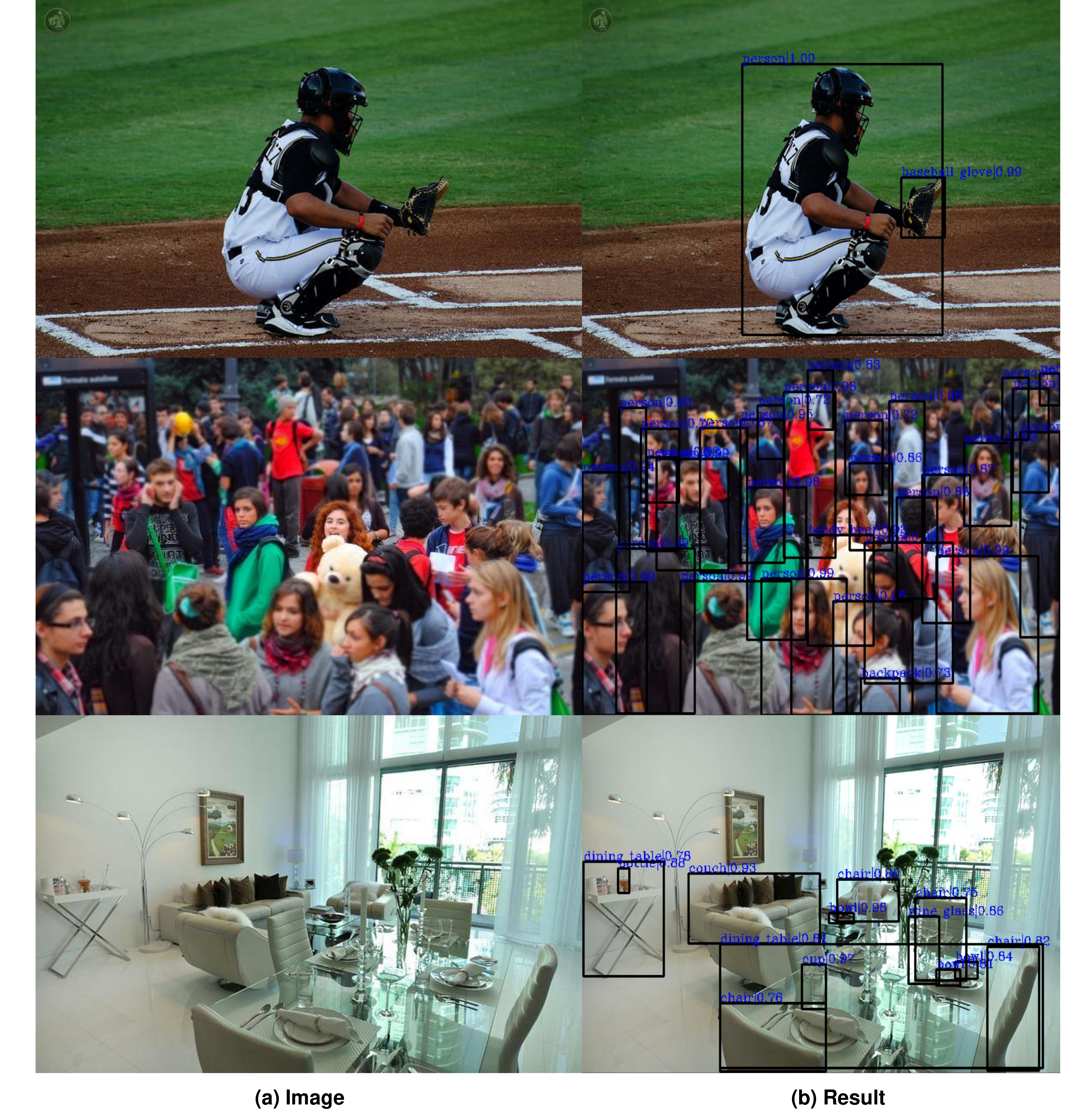}
	\caption{Sample detection results from our GraphFPN based method. Images are sampled from the MS COCO 2017 validation set~\cite{lin2014microsoft}.}
	\label{Fig:res-visual2}
\end{figure*}

\begin{figure*}[t]
	\centering
	\includegraphics[width=1.0\linewidth]{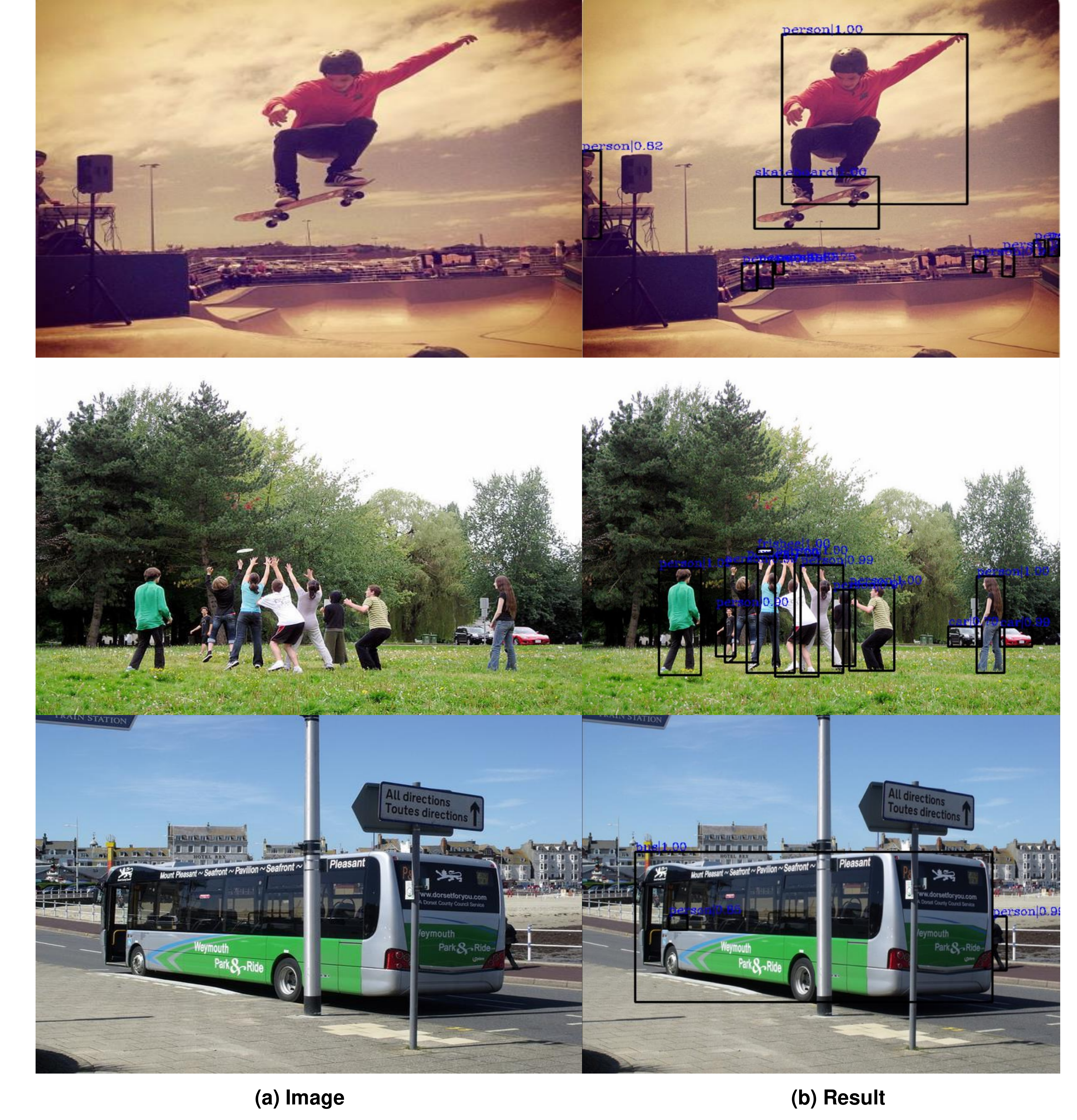}
	\caption{Sample detection results from our GraphFPN based method. Images are sampled from the MS COCO 2017 validation set~\cite{lin2014microsoft}.}
	\label{Fig:res-visual3}
\end{figure*}

Figure~\ref{Fig:res-visual1} shows sample detection results from FPN~\cite{lin2017feature}, FPT~\cite{zhang2020feature}, and our GraphFPN based method. Input images are taken from the MS COCO 2017 validation set~\cite{lin2014microsoft}.
Figures~\ref{Fig:res-visual2} and~\ref{Fig:res-visual3} show additional sample detection results from our GraphFPN based method. Images are taken from the MS COCO 2017 validation set~\cite{lin2014microsoft}.

\section{Experiments on Semantic Segmentation}
To demonstrate the effectiveness of our method in capturing intrinsic image structures, we further apply our method to semantic segmentation. In our experiments in semantic segmentation, we test the performance of UFP + GraphFPN and compare its results with unscathed feature pyramid networks(UFP~\cite{peng2017large}) and feature pyramid transformer. Table~\ref{tab:cityscapes} shows experimental results on the Cityscapes~\cite{cordts2016cityscapes} dataset, which contains 19 classes and includes 2,975,500 images for training and validation. The settings of this experiment are the same as in~\cite{zhang2020feature}. We also adopt Unscathed Feature Pyramid (UFP)~\cite{peng2017large} as the feature pyramid construction module. From the experimental results shown in Table~\ref{tab:cityscapes}, it can be found out that our proposed method achieves clearly better performance, which also demonstrates the applicability of our method.

\begin{table*}[t]
\begin{center}
\begin{tabular}{c|l|c|c}
\toprule[1pt]
 Stage & Layer Name & Output Size & Kernels, \#channels   \\
 \hline
$C_{1}$ & conv1 & $W \times H$ & 7 $\times$ 7, 64,stride 2    \\
\hline
 \multirow{2}{*}{$C_{2}$} &  \multirow{2}{*}{conv2\_x} &  \multirow{2}{*}{$\frac{W}{2}$ $\times$ $\frac{H}{2}$} & 3 $\times$ 3 max pool, stride 2  \\
 & & & $\begin{bmatrix} 1 \times 1, 64 \\ 3 \times 3, 64  \\  1 \times 1, 256  \end{bmatrix} \times 3$\\
\hline
$C_{3}$ & conv3\_x & $\frac{W}{4}$ $\times$ $\frac{H}{4}$  & $\begin{bmatrix} 1 \times 1, 128 \\ 3 \times 3, 128  \\  1 \times 1, 512  \end{bmatrix} \times 23$\\
\hline
 $C_{4}$ & conv4\_x & $\frac{W}{8}$ $\times$ $\frac{H}{8}$  & $\begin{bmatrix} 1 \times 1, 256 \\ 3 \times 3, 256  \\  1 \times 1, 1024  \end{bmatrix} \times 4$ \\
\hline 
 $C_{5}$ & conv5\_x & $\frac{W}{16}$ $\times$ $\frac{H}{16}$  & $\begin{bmatrix} 1 \times 1, 512 \\ 3 \times 3, 512  \\  1 \times 1, 2048  \end{bmatrix} \times 3$\\
\hline 
 $P_{1}$ & - & $W$ $\times$  $H$  &  3 $\times$ 3, 256 \\
\hline
 $P_{2}$ & - & $\frac{W}{2}$ $\times$ $\frac{H}{2}$ & 3 $\times$ 3, 256   \\
\hline 
 $P_{3}$ & - & $\frac{W}{4}$ $\times$ $\frac{H}{4}$ & 3 $\times$ 3, 256  \\
\hline 
 $P_{4}$ & - & $\frac{W}{8}$ $\times$ $\frac{H}{8}$ & 3 $\times$ 3, 256    \\
\hline
 $P_{5}$ & - & $\frac{W}{16}$ $\times$ $\frac{H}{16}$ & 3 $\times$ 3, 256   \\
\bottomrule[1pt]
\end{tabular}
\end{center}
\caption{Network architecture of the backbone (ResNet-101) and convolutional FPN used in the proposed pipeline. Residual building blocks are shown in brackets, with the numbers of blocks stacked. Downsampling is performed by conv3\_1, conv4\_1, and conv5\_1 with a stride of 2. $W$ and $H$ are the input width and height.}\label{tab:backbone+FPN}
\end{table*}

\begin{table*}[t]
\begin{center}
\begin{tabular}{c|l|c|c}
\toprule[1pt]
 Stage & Layer Name & \#Node & \#Feature Channel  \\
 \hline
\multirow{3}{*}{CGL-1} & CL-1 & \multirow{3}{*}{$N+\frac{N}{4}+\frac{N}{16}+\frac{N}{64}+\frac{N}{256}$} & \multirow{3}{*}{256}     \\
& CL-2 &  &     \\
& CL-3 &  &     \\
\midrule
\multirow{3}{*}{HGL} & HL-1 & \multirow{3}{*}{$N+\frac{N}{4}+\frac{N}{16}+\frac{N}{64}+\frac{N}{256}$} & \multirow{3}{*}{256}     \\
& HL-2 &  &     \\
& HL-3 &  &     \\
\midrule
\multirow{3}{*}{CGL-2} & CL-4 & \multirow{3}{*}{$N+\frac{N}{4}+\frac{N}{16}+\frac{N}{64}+\frac{N}{256}$} & \multirow{3}{*}{256}     \\
& CL-5 &  &     \\
& CL-6 &  &     \\
\bottomrule[1pt]
\end{tabular}
\end{center}
\caption{Network architecture of our GraphFPN. ``CGL-1'' stands for the first group of contextual layers, ``HGL'' stands for the group of hierarchical layers, and ``CGL-2'' stands for the second group of contextual layers. Each ``CL'' or ``HL'' layer has three attention modules. Note that the number of graph nodes in each layer is $N+\frac{N}{4}+\frac{N}{16}+\frac{N}{64}+\frac{N}{256}$, where $N$ is the number of superpixels in the finest level of a superpixel hierarchy. }\label{tab:graphFPN}
\end{table*}

\begin{table}[t]
\begin{center}

\resizebox{0.5\linewidth}{!}{

\begin{tabular}{c|cccc}
\toprule[1pt]
Methods & Train.mIoU & Val.mIoU & Params           & GFLOPs  \\
\midrule
UFP~\cite{peng2017large} & 86.0 & 79.1 & 71.3 M & 916.1  \\
UFP+FPS~\cite{zhang2020feature} & 87.4 & 81.7 & 127.2 M & 1063.9  \\ \midrule
UFP+GraphFPN &  \textbf{88.4} ($\uparrow$1.0) & \textbf{83.2}($\uparrow$1.5) & 130.1 M & 1104.2  \\
\bottomrule[1pt]
\end{tabular}
}
\end{center}
\caption{Comparison with state-of-the-art semantic segmentation methods on the Cityscapes validation set~\cite{cordts2016cityscapes}.}\label{tab:cityscapes}
\end{table}

%% file: egpaper_final.bbl
\begin{thebibliography}{10}\itemsep=-1pt

\bibitem{2019Heterogeneous}
Heterogeneous graph attention network.
\newblock In {\em The World Wide Web Conference}, 2019.

\bibitem{belongie2002shape}
Serge Belongie, Jitendra Malik, and Jan Puzicha.
\newblock Shape matching and object recognition using shape contexts.
\newblock {\em IEEE transactions on pattern analysis and machine intelligence},
  24(4):509--522, 2002.

\bibitem{bienenstock1997compositionality}
Elie Bienenstock, Stuart Geman, and Daniel Potter.
\newblock Compositionality, mdl priors, and object recognition.
\newblock {\em Advances in neural information processing systems}, pages
  838--844, 1997.

\bibitem{bilinski2018dense}
Piotr Bilinski and Victor Prisacariu.
\newblock Dense decoder shortcut connections for single-pass semantic
  segmentation.
\newblock In {\em Proceedings of the IEEE Conference on Computer Vision and
  Pattern Recognition}, pages 6596--6605, 2018.

\bibitem{carion2020end}
Nicolas Carion, Francisco Massa, Gabriel Synnaeve, Nicolas Usunier, Alexander
  Kirillov, and Sergey Zagoruyko.
\newblock End-to-end object detection with transformers.
\newblock In {\em European Conference on Computer Vision}, pages 213--229.
  Springer, 2020.

\bibitem{cordts2016cityscapes}
Marius Cordts, Mohamed Omran, Sebastian Ramos, Timo Rehfeld, Markus Enzweiler,
  Rodrigo Benenson, Uwe Franke, Stefan Roth, and Bernt Schiele.
\newblock The cityscapes dataset for semantic urban scene understanding.
\newblock In {\em Proceedings of the IEEE conference on computer vision and
  pattern recognition}, pages 3213--3223, 2016.

\bibitem{fu2019dual}
Jun Fu, Jing Liu, Haijie Tian, Yong Li, Yongjun Bao, Zhiwei Fang, and Hanqing
  Lu.
\newblock Dual attention network for scene segmentation.
\newblock In {\em Proceedings of the IEEE/CVF Conference on Computer Vision and
  Pattern Recognition}, pages 3146--3154, 2019.

\bibitem{gao2019graph}
Hongyang Gao and Shuiwang Ji.
\newblock Graph u-nets.
\newblock In {\em Proceedings of the 36th International Conference on Machine
  Learning}, 2019.

\bibitem{2019NAS}
G. Ghiasi, T.~Y. Lin, and Q.~V. Le.
\newblock Nas-fpn: Learning scalable feature pyramid architecture for object
  detection.
\newblock In {\em 2019 IEEE/CVF Conference on Computer Vision and Pattern
  Recognition (CVPR)}, 2019.

\bibitem{girshick2014rich}
Ross Girshick, Jeff Donahue, Trevor Darrell, and Jitendra Malik.
\newblock Rich feature hierarchies for accurate object detection and semantic
  segmentation.
\newblock In {\em Proceedings of the IEEE conference on computer vision and
  pattern recognition}, pages 580--587, 2014.

\bibitem{Gong_2019_CVPR}
Liyu Gong and Qiang Cheng.
\newblock Exploiting edge features for graph neural networks.
\newblock In {\em The IEEE Conference on Computer Vision and Pattern
  Recognition (CVPR)}, June 2019.

\bibitem{2020AugFPN}
C. Guo, B. Fan, Q. Zhang, S. Xiang, and C. Pan.
\newblock Augfpn: Improving multi-scale feature learning for object detection.
\newblock In {\em 2020 IEEE/CVF Conference on Computer Vision and Pattern
  Recognition (CVPR)}, 2020.

\bibitem{he2016deep}
Kaiming He, Xiangyu Zhang, Shaoqing Ren, and Jian Sun.
\newblock Deep residual learning for image recognition.
\newblock In {\em Proceedings of the IEEE conference on computer vision and
  pattern recognition}, pages 770--778, 2016.

\bibitem{hinton1979some}
Geoffrey Hinton.
\newblock Some demonstrations of the effects of structural descriptions in
  mental imagery.
\newblock {\em Cognitive Science}, 3(3):231--250, 1979.

\bibitem{hinton2021represent}
Geoffrey Hinton.
\newblock How to represent part-whole hierarchies in a neural network.
\newblock {\em arXiv preprint arXiv:2102.12627}, 2021.

\bibitem{hinton2018matrix}
Geoffrey~E Hinton, Sara Sabour, and Nicholas Frosst.
\newblock Matrix capsules with em routing.
\newblock In {\em International conference on learning representations}, 2018.

\bibitem{hu2018squeeze}
Jie Hu, Li Shen, and Gang Sun.
\newblock Squeeze-and-excitation networks.
\newblock In {\em Proceedings of the IEEE conference on computer vision and
  pattern recognition}, pages 7132--7141, 2018.

\bibitem{jha2020doubleu}
Debesh Jha, Michael~A Riegler, Dag Johansen, P{\aa}l Halvorsen, and
  H{\aa}vard~D Johansen.
\newblock Doubleu-net: A deep convolutional neural network for medical image
  segmentation.
\newblock In {\em 2020 IEEE 33rd International Symposium on Computer-Based
  Medical Systems (CBMS)}, pages 558--564. IEEE, 2020.

\bibitem{2018SAN}
Y. Kim, B.~N. Kang, and D. Kim.
\newblock San: Learning relationship between convolutional features for
  multi-scale object detection: 15th european conference, munich, germany,
  september 8–14, 2018, proceedings, part v.
\newblock In {\em Springer, Cham}, 2018.

\bibitem{kingma2014adam}
Diederik~P Kingma and Jimmy Ba.
\newblock Adam: A method for stochastic optimization.
\newblock {\em arXiv preprint arXiv:1412.6980}, 2014.

\bibitem{kipf2016semi}
Thomas~N Kipf and Max Welling.
\newblock Semi-supervised classification with graph convolutional networks.
\newblock {\em arXiv preprint arXiv:1609.02907}, 2016.

\bibitem{2018Deep}
T. Kong, F. Sun, W. Huang, and H. Liu.
\newblock Deep feature pyramid reconfiguration for object detection.
\newblock 2018.

\bibitem{kosiorek2019stacked}
Adam~R Kosiorek, Sara Sabour, Yee~Whye Teh, and Geoffrey~E Hinton.
\newblock Stacked capsule autoencoders.
\newblock {\em arXiv preprint arXiv:1906.06818}, 2019.

\bibitem{krizhevsky2012imagenet}
Alex Krizhevsky, Ilya Sutskever, and Geoffrey~E Hinton.
\newblock Imagenet classification with deep convolutional neural networks.
\newblock In {\em Advances in neural information processing systems}, pages
  1097--1105, 2012.

\bibitem{li2019deepgcns}
Guohao Li, Matthias Muller, Ali Thabet, and Bernard Ghanem.
\newblock Deepgcns: Can gcns go as deep as cnns?
\newblock In {\em Proceedings of the IEEE International Conference on Computer
  Vision}, pages 9267--9276, 2019.

\bibitem{li2019scale}
Yanghao Li, Yuntao Chen, Naiyan Wang, and Zhaoxiang Zhang.
\newblock Scale-aware trident networks for object detection.
\newblock In {\em Proceedings of the IEEE/CVF International Conference on
  Computer Vision}, pages 6054--6063, 2019.

\bibitem{li2018detnet}
Zeming Li, Chao Peng, Gang Yu, Xiangyu Zhang, Yangdong Deng, and Jian Sun.
\newblock Detnet: Design backbone for object detection.
\newblock In {\em Proceedings of the European conference on computer vision
  (ECCV)}, pages 334--350, 2018.

\bibitem{lin2018multi}
Di Lin, Yuanfeng Ji, Dani Lischinski, Daniel Cohen-Or, and Hui Huang.
\newblock Multi-scale context intertwining for semantic segmentation.
\newblock In {\em Proceedings of the European Conference on Computer Vision
  (ECCV)}, pages 603--619, 2018.

\bibitem{lin2019zigzagnet}
Di Lin, Dingguo Shen, Siting Shen, Yuanfeng Ji, Dani Lischinski, Daniel
  Cohen-Or, and Hui Huang.
\newblock Zigzagnet: Fusing top-down and bottom-up context for object
  segmentation.
\newblock In {\em Proceedings of the IEEE/CVF Conference on Computer Vision and
  Pattern Recognition}, pages 7490--7499, 2019.

\bibitem{lin2017feature}
Tsung-Yi Lin, Piotr Doll{\'a}r, Ross Girshick, Kaiming He, Bharath Hariharan,
  and Serge Belongie.
\newblock Feature pyramid networks for object detection.
\newblock In {\em Proceedings of the IEEE conference on computer vision and
  pattern recognition}, pages 2117--2125, 2017.

\bibitem{2017Focal}
T.~Y. Lin, P. Goyal, R. Girshick, K. He, and P Dollár.
\newblock Focal loss for dense object detection.
\newblock {\em IEEE Transactions on Pattern Analysis and Machine Intelligence},
  PP(99):2999--3007, 2017.

\bibitem{lin2014microsoft}
Tsung-Yi Lin, Michael Maire, Serge Belongie, James Hays, Pietro Perona, Deva
  Ramanan, Piotr Doll{\'a}r, and C~Lawrence Zitnick.
\newblock Microsoft coco: Common objects in context.
\newblock In {\em European conference on computer vision}, pages 740--755.
  Springer, 2014.

\bibitem{liu2018path}
Shu Liu, Lu Qi, Haifang Qin, Jianping Shi, and Jiaya Jia.
\newblock Path aggregation network for instance segmentation.
\newblock In {\em Proceedings of the IEEE conference on computer vision and
  pattern recognition}, pages 8759--8768, 2018.

\bibitem{liu2016ssd}
Wei Liu, Dragomir Anguelov, Dumitru Erhan, Christian Szegedy, Scott Reed,
  Cheng-Yang Fu, and Alexander~C Berg.
\newblock Ssd: Single shot multibox detector.
\newblock In {\em European conference on computer vision}, pages 21--37.
  Springer, 2016.

\bibitem{maninis2017convolutional}
Kevis-Kokitsi Maninis, Jordi Pont-Tuset, Pablo Arbel{\'a}ez, and Luc Van~Gool.
\newblock Convolutional oriented boundaries: From image segmentation to
  high-level tasks.
\newblock {\em IEEE transactions on pattern analysis and machine intelligence},
  40(4):819--833, 2017.

\bibitem{marr1982vision}
David Marr.
\newblock Vision: A computational investigation into the human representation
  and processing of visual information.
\newblock 1982.

\bibitem{pantofaru2008object}
Caroline Pantofaru, Cordelia Schmid, and Martial Hebert.
\newblock Object recognition by integrating multiple image segmentations.
\newblock In {\em European conference on computer vision}, pages 481--494.
  Springer, 2008.

\bibitem{peng2018megdet}
Chao Peng, Tete Xiao, Zeming Li, Yuning Jiang, Xiangyu Zhang, Kai Jia, Gang Yu,
  and Jian Sun.
\newblock Megdet: A large mini-batch object detector.
\newblock In {\em Proceedings of the IEEE Conference on Computer Vision and
  Pattern Recognition}, pages 6181--6189, 2018.

\bibitem{peng2017large}
Chao Peng, Xiangyu Zhang, Gang Yu, Guiming Luo, and Jian Sun.
\newblock Large kernel matters--improve semantic segmentation by global
  convolutional network.
\newblock In {\em Proceedings of the IEEE conference on computer vision and
  pattern recognition}, pages 4353--4361, 2017.

\bibitem{pont2016multiscale}
Jordi Pont-Tuset, Pablo Arbelaez, Jonathan~T Barron, Ferran Marques, and
  Jitendra Malik.
\newblock Multiscale combinatorial grouping for image segmentation and object
  proposal generation.
\newblock {\em IEEE transactions on pattern analysis and machine intelligence},
  39(1):128--140, 2016.

\bibitem{redmon2016you}
Joseph Redmon, Santosh Divvala, Ross Girshick, and Ali Farhadi.
\newblock You only look once: Unified, real-time object detection.
\newblock In {\em Proceedings of the IEEE conference on computer vision and
  pattern recognition}, pages 779--788, 2016.

\bibitem{ren2015faster}
Shaoqing Ren, Kaiming He, Ross Girshick, and Jian Sun.
\newblock Faster r-cnn: Towards real-time object detection with region proposal
  networks.
\newblock In {\em Advances in neural information processing systems}, pages
  91--99, 2015.

\bibitem{ren2016faster}
Shaoqing Ren, Kaiming He, Ross Girshick, and Jian Sun.
\newblock Faster r-cnn: towards real-time object detection with region proposal
  networks.
\newblock {\em IEEE transactions on pattern analysis and machine intelligence},
  39(6):1137--1149, 2016.

\bibitem{ronneberger2015u}
Olaf Ronneberger, Philipp Fischer, and Thomas Brox.
\newblock U-net: Convolutional networks for biomedical image segmentation.
\newblock In {\em International Conference on Medical image computing and
  computer-assisted intervention}, pages 234--241. Springer, 2015.

\bibitem{sabour2017dynamic}
Sara Sabour, Nicholas Frosst, and Geoffrey~E Hinton.
\newblock Dynamic routing between capsules.
\newblock {\em arXiv preprint arXiv:1710.09829}, 2017.

\bibitem{shrivastava2016beyond}
Abhinav Shrivastava, Rahul Sukthankar, Jitendra Malik, and Abhinav Gupta.
\newblock Beyond skip connections: Top-down modulation for object detection.
\newblock {\em arXiv preprint arXiv:1612.06851}, 2016.

\bibitem{2020Sparse}
P. Sun, R. Zhang, Y. Jiang, T. Kong, C. Xu, W. Zhan, M. Tomizuka, L. Li, Z.
  Yuan, and C. Wang.
\newblock Sparse r-cnn: End-to-end object detection with learnable proposals.
\newblock 2020.

\bibitem{szegedy2015going}
Christian Szegedy, Wei Liu, Yangqing Jia, Pierre Sermanet, Scott Reed, Dragomir
  Anguelov, Dumitru Erhan, Vincent Vanhoucke, and Andrew Rabinovich.
\newblock Going deeper with convolutions.
\newblock In {\em Proceedings of the IEEE conference on computer vision and
  pattern recognition}, pages 1--9, 2015.

\bibitem{2019EfficientNet}
M. Tan and Q.~V. Le.
\newblock Efficientnet: Rethinking model scaling for convolutional neural
  networks.
\newblock 2019.

\bibitem{tao2020hierarchical}
Andrew Tao, Karan Sapra, and Bryan Catanzaro.
\newblock Hierarchical multi-scale attention for semantic segmentation.
\newblock {\em arXiv preprint arXiv:2005.10821}, 2020.

\bibitem{velivckovic2017graph}
Petar Veli{\v{c}}kovi{\'c}, Guillem Cucurull, Arantxa Casanova, Adriana Romero,
  Pietro Lio, and Yoshua Bengio.
\newblock Graph attention networks.
\newblock {\em ICLR}, 2018.

\bibitem{2020Auto}
H. Xu, L. Yao, Z. Li, X. Liang, and W. Zhang.
\newblock Auto-fpn: Automatic network architecture adaptation for object
  detection beyond classification.
\newblock In {\em 2019 IEEE/CVF International Conference on Computer Vision
  (ICCV)}, 2020.

\bibitem{2018Hierarchical}
R. Ying, J. You, C. Morris, X. Ren, William~L Hamilton, and J. Leskovec.
\newblock Hierarchical graph representation learning with differentiable
  pooling.
\newblock 2018.

\bibitem{zhang2020feature}
Dong Zhang, Hanwang Zhang, Jinhui Tang, Meng Wang, Xiansheng Hua, and Qianru
  Sun.
\newblock Feature pyramid transformer.
\newblock In {\em European Conference on Computer Vision}, pages 323--339.
  Springer, 2020.

\bibitem{zhang2018context}
Hang Zhang, Kristin Dana, Jianping Shi, Zhongyue Zhang, Xiaogang Wang, Ambrish
  Tyagi, and Amit Agrawal.
\newblock Context encoding for semantic segmentation.
\newblock In {\em Proceedings of the IEEE conference on Computer Vision and
  Pattern Recognition}, pages 7151--7160, 2018.

\bibitem{zhang2018exfuse}
Zhenli Zhang, Xiangyu Zhang, Chao Peng, Xiangyang Xue, and Jian Sun.
\newblock Exfuse: Enhancing feature fusion for semantic segmentation.
\newblock In {\em Proceedings of the European Conference on Computer Vision
  (ECCV)}, pages 269--284, 2018.

\bibitem{zhao2019m2det}
Qijie Zhao, Tao Sheng, Yongtao Wang, Zhi Tang, Ying Chen, Ling Cai, and Haibin
  Ling.
\newblock M2det: A single-shot object detector based on multi-level feature
  pyramid network.
\newblock In {\em Proceedings of the AAAI conference on artificial
  intelligence}, volume~33, pages 9259--9266, 2019.

\bibitem{zhu2020deformable}
Xizhou Zhu, Weijie Su, Lewei Lu, Bin Li, Xiaogang Wang, and Jifeng Dai.
\newblock Deformable detr: Deformable transformers for end-to-end object
  detection.
\newblock {\em arXiv preprint arXiv:2010.04159}, 2020.

\end{thebibliography}
